\documentclass[11pt]{article}
\usepackage{coling2020}
\usepackage{times}
\usepackage{url}
\usepackage{latexsym}

\usepackage{microtype}
\usepackage{graphicx}
\usepackage{booktabs}
\usepackage{numprint}
\usepackage{amssymb}
\usepackage{multirow}
\usepackage{todonotes} 
\usepackage{xspace}
\usepackage{stfloats}
\usepackage{comment}
\usepackage{hyperref}

\colingfinalcopy % Uncomment this line for the final submission

\title{CharacterBERT: Reconciling ELMo and BERT for Word-Level Open-Vocabulary Representations From Characters}

\author{
  Hicham El Boukkouri$^{\mathbf{1}}$, Olivier Ferret$^{\mathbf{2}}$, Thomas Lavergne$^{\mathbf{1}}$, Hiroshi Noji$^{\mathbf{3}}$,\\ \textbf{Pierre Zweigenbaum}$^{\mathbf{1}}$, \textbf{Junichi Tsujii}$^{\mathbf{3}}$ \\[6pt]
  $^1$Universit\'e Paris-Saclay, CNRS, LIMSI, Orsay, France,\\
  $^2$Universit\'e Paris-Saclay, CEA, List, F-91120, Palaiseau, France, \\
  $^3$Artificial Intelligence Research Center (AIRC), AIST, Japan \\
  \texttt{\{elboukkouri,lavergne,pz\}@limsi.fr},  \texttt{olivier.ferret@cea.fr}\\ 
  \texttt{\{hiroshi.noji,j-tsujii\}@aist.go.jp}\\ 
  }

\date{}

\begin{document}
\maketitle

\begin{abstract}
Due to the compelling improvements brought by BERT, many recent representation models adopted the Transformer architecture as their main building block, consequently inheriting the \textit{wordpiece} tokenization system despite it not being intrinsically linked to the notion of Transformers. While this system is thought to achieve a good balance between the flexibility of characters and the efficiency of full words, using predefined wordpiece vocabularies from the general domain is not always suitable, especially when building models for \textit{specialized domains} (e.g., the medical domain). Moreover, adopting a wordpiece tokenization shifts the focus from the word level to the subword level, making the models conceptually more complex and arguably less convenient in practice. For these reasons, we propose CharacterBERT, a new variant of BERT that drops the wordpiece system altogether and uses a Character-CNN module instead to represent \textit{entire words} by consulting their \textit{characters}. We show that this new model improves the performance of BERT on a variety of medical domain tasks while at the same time producing robust, word-level, and open-vocabulary representations. 
\end{abstract}

\section{Introduction}
Pre-trained language representations from Transformers \cite{vaswani2017attention} have become arguably the most popular choice for building NLP systems\footnote{See the leaderboard of the \href{https://gluebenchmark.com/leaderboard}{GLUE benchmark}.}. Among all such models, BERT \cite{devlin-etal-2019-bert} has probably been the most successful, spawning a large number of new improved variants \cite{liu2019roberta,lan2019albert,sun2019ernie,zhang2019ernie,clark2020electra}. As a result, many of the recent language representation models inherited BERT's subword tokenization system which relies on a predefined set of \textit{wordpieces} \cite{wu2016google}, supposedly striking a good balance between the flexibility of characters and the efficiency of full words.

While current research mostly focuses on improving language representations for the default ``general-domain'', there seems to be a growing interest in building suitable word embeddings for more \textit{specialized} domains \cite{el-boukkouri-etal-2019-embedding,si2019enhancing,elwany2019bert}. However, with the growing complexity of recent representation models, the default trend seems to favor re-training general-domain models on specialized corpora rather than building models from scratch with a specialized vocabulary (e.g., BlueBERT \cite{peng2019transfer} and BioBERT \cite{lee2020biobert}). While these methods undeniably produce good models \footnote{See the baselines from the \href{https://github.com/ncbi-nlp/BLUE_Benchmark/\#baselines}{BLUE benchmark}.}, a few questions remain: How suitable are the predefined general-domain vocabularies when used in the context of specialized domains (e.g., the medical domain)? Is it better to train specialized models with specialized subword units? Do we induce any biases by training specialized models with general-domain wordpieces?

In this paper, we propose CharacterBERT, a possible solution for avoiding any biases that may come from the use of a predefined wordpiece vocabulary, and an effort to revert back to conceptually simpler word-level models. This new variant does not rely on wordpieces but instead consults the characters of each token to build representations similarly to previous word-level open-vocabulary systems \cite{luong-manning-2016-achieving,kim2016character,jozefowicz2016exploring}. In practice, we replace BERT's wordpiece embedding layer with ELMo's \cite{peters-etal-2018-deep} Character-CNN module while keeping the rest of the architecture untouched. As a result, CharacterBERT is able to produce word-level contextualized representations and does not require a wordpiece vocabulary. Furthermore, this new model seems better suited than vanilla BERT for training specialized models, as evidenced by an evaluation on multiple tasks from the medical domain. Finally, as expected from a character-based system, CharacterBERT is also seemingly more robust to noise and misspellings. To the best of our knowledge, this is the first work that replaces BERT's wordpiece system with a word-level character-based system. \\

Our contributions are the following:
\begin{itemize}
    \item We provide preliminary evidence that general-domain wordpiece vocabularies are not suitable for specialized domain applications.
    \item We propose CharacterBERT, a new variant of BERT that produces word-level contextual representations by consulting characters. \item We evaluate CharacterBERT on multiple specialized medical tasks and show that it outperforms BERT without requiring a wordpiece vocabulary.
    \item We exhibit signs of improved robustness to noise and misspellings in favor of CharacterBERT.
    \item We enable the reproducibility of our experiments by sharing our pre-training and fine-tuning codes. Furthermore, we also share our pre-trained representation models to benefit the NLP community\footnote{Our models and code are available at: \url{https://github.com/helboukkouri/character-bert}.}.
\end{itemize}

This work has only focused on the English language and the medical (clinical and biomedical) domain. The generalization to other languages and specialized domains is left to future work.

\section{General-Domain Wordpieces in Specialized Domains}

Since many specialized versions of BERT come from re-training the original model on a set of specialized texts, we carry out a couple of preliminary experiments to gauge the effect of using a general-domain wordpiece vocabulary in a specialized domain. Here we focus on the medical domain for which we learn\footnote{We use the open-source implementation in \href{https://github.com/google/sentencepiece}{SentencePiece}.} a new wordpiece vocabulary using MIMIC-III clinical notes \cite{johnson2016mimic} and PMC OA\footnote{\href{https://www.ncbi.nlm.nih.gov/pmc/tools/openftlist/}{PubMed Central Open Access}.} biomedical article abstracts. We then process a sample (1M tokens) of the medical corpus with either the medical vocabulary or BERT's original vocabulary and examine the difference.

\begin{figure}[htbp]
\begin{center} 
\includegraphics[height=160pt]{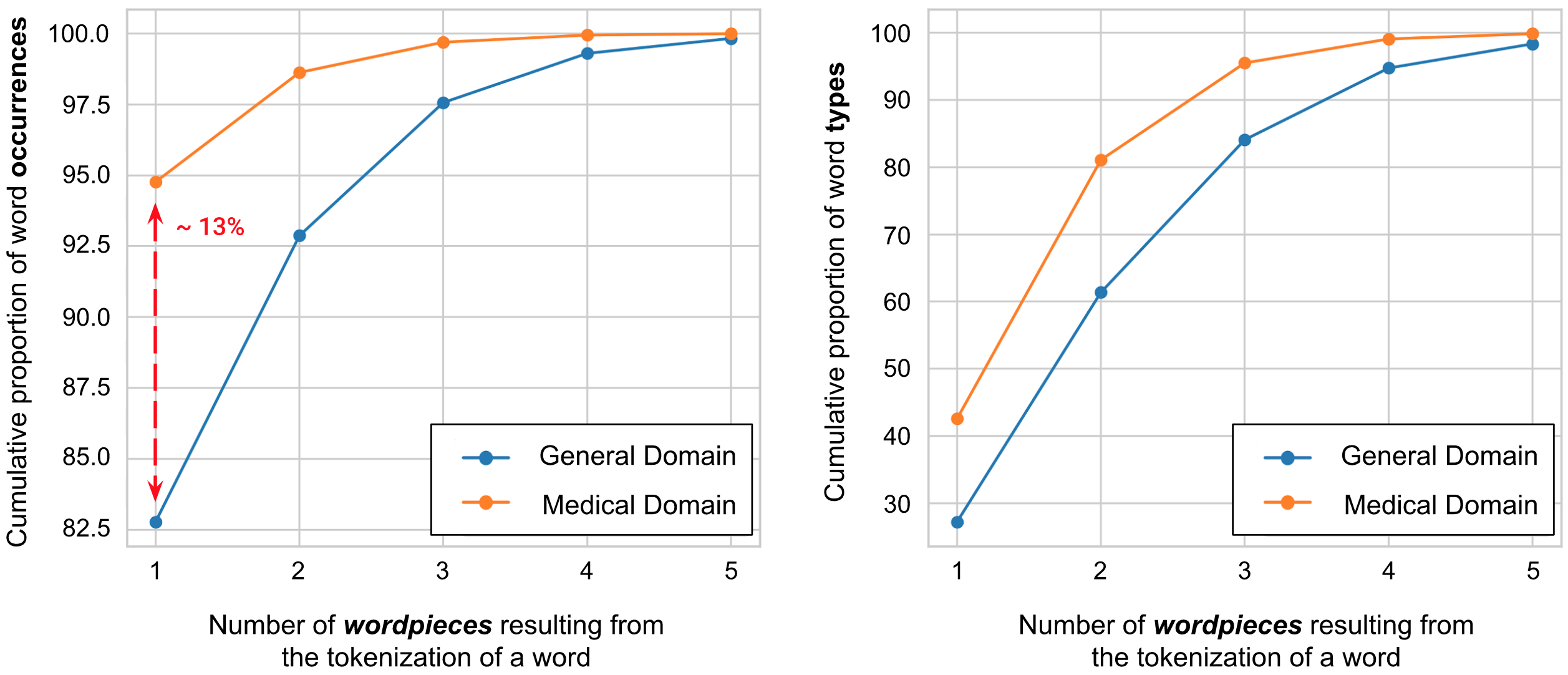}
\end{center} 
\caption{Comparison of the tokenization of a medical corpus by vocabularies from different domains.} \label{vocabCompare1}
\end{figure}

Looking at the frequency of splitting an unknown token into multiple wordpieces (cf.~Figure~\ref{vocabCompare1}) we see that the medical vocabulary produces overall less wordpieces than the general version, both at occurrence and type levels. Moreover, we see that $\approx 13\%$ of occurrences are never split as they are already part of the medical vocabulary but are decomposed into two or more wordpieces by the general vocabulary.

\begin{table}[htbp]
\centering
\setlength{\tabcolsep}{3pt}
\resizebox{0.6\textwidth}{!}{%
\begin{tabular}{lll}
\toprule
\textbf{Reference\qquad} & \textbf{Medical Vocabulary} & \textbf{General Vocabulary}\\
\midrule
paracetamol & [paracetamol & [para, ce, tam, ol] \\
choledocholithiasis & [choledoch, olithiasis] & [cho, led, och, oli, thi, asi, s] \\
borborygmi & [bor, bor, yg, mi] & [bo, rb, ory, gm, i] \\
\bottomrule
\end{tabular}
}
\centering\caption{Comparison of the tokenization of specific medical terms by vocabularies from different domains.}\label{vocabCompare2}
\end{table}

When looking closer at the quality of the produced wordpieces (cf. Table~\ref{vocabCompare2}), we see that in addition to producing fewer subwords, the specialized vocabulary also seems to produce more meaningful units (e.g. ``choledoch'' and ``olithiasis''). These preliminary analyses show that the choice of a vocabulary affects the quality of the tokenization which may in turn induce biases in downstream applications of the representation model. To avoid such biases, and in an effort to revert back to more convenient and conceptually simpler word-level models, we propose CharacterBERT, a wordpiece-free variant of BERT.

\section{CharacterBERT}

\begin{figure}[htbp] 
\begin{center} 
\includegraphics[height=270pt]{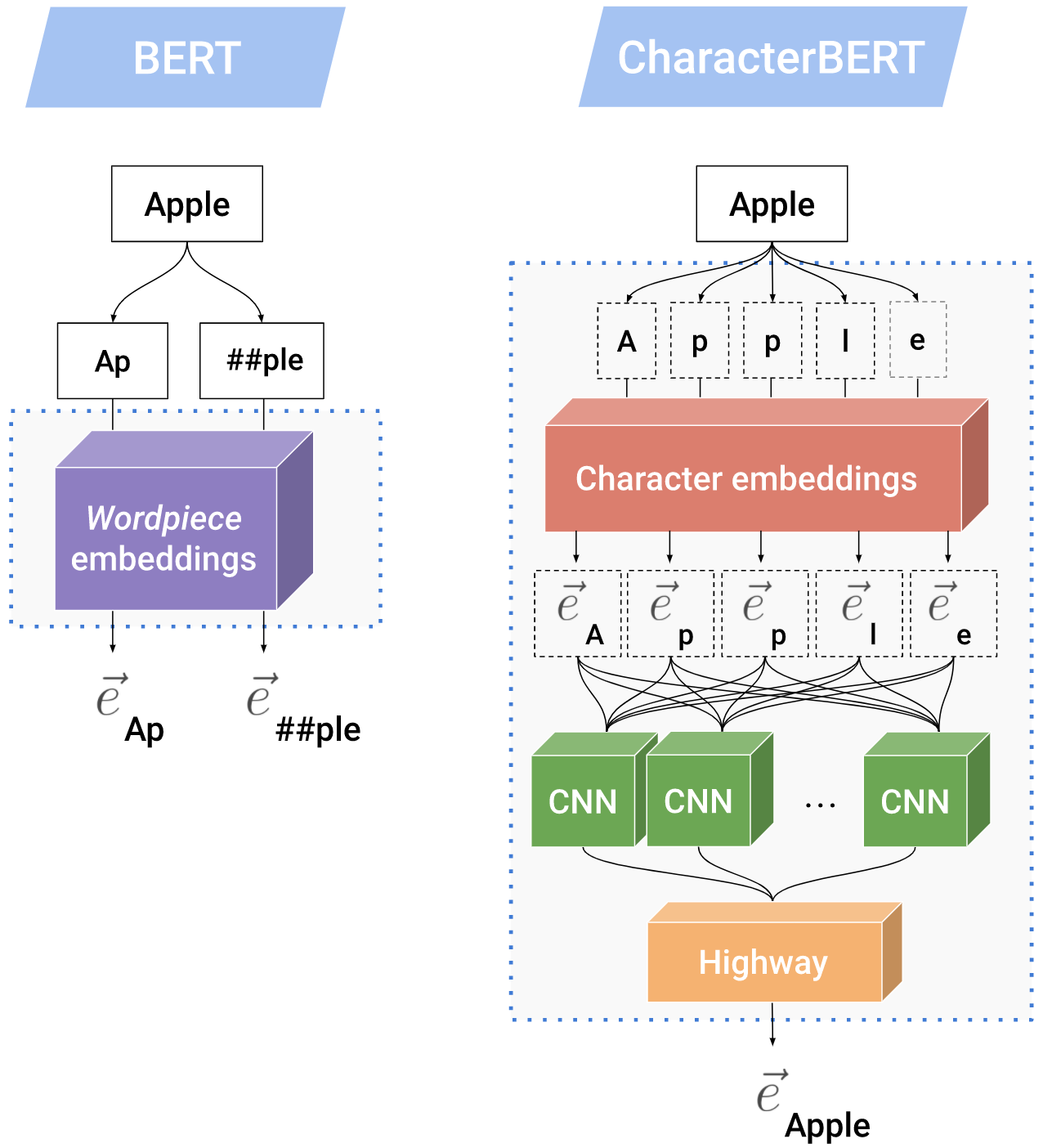}
\end{center} 
\caption{Comparison of the context-independent representation systems in BERT and CharacterBERT. In this illustration, BERT splits the word ``Apple'' into two wordpieces then embeds each unit separately. CharacterBERT produces a single embedding for ``Apple'' by consulting its sequence of characters.} \label{bertvscharacterbert}
\end{figure}

CharacterBERT is similar in every way to vanilla BERT but uses a different method to construct initial context-independent representations: while the original model consults its vocabulary to split unknown tokens into multiple wordpieces then embeds each unit independently using a wordpiece embedding matrix, CharacterBERT uses a Character-CNN module \cite{peters-etal-2018-deep,jozefowicz2016exploring} which consults the characters of a token to produce a single representation (see Figure~\ref{bertvscharacterbert}).

\subsection{Character-CNN: Building Word Representations From Characters}

We use the Character-CNN that is implemented as part of ELMo's architecture. This module constructs context-independent token representations through the following steps:
\begin{enumerate}
    \item Each token is converted into a sequence of characters\footnote{In practice, the tokens are encoded in UTF-8 and all characters including non-ascii symbols are converted into bytes. This allows us to keep a small byte vocabulary of size 256 to which we add a few special symbols for a total of 263.} with a maximum sequence length of 50. \item A lookup is performed for each character, producing a sequence of 16-d embeddings. \item The character embedding sequence is fed to multiple 1-d CNNs \cite{lecun1989backpropagation} with different filters\footnote{We use seven 1-d CNNs with the following filters: [1, 32], [2, 32], [3, 64], [4, 128], [5, 256], [6, 512] and [7, 1024].}. The output of each CNN is then max-pooled across the character sequence and concatenated with other CNN outputs to produce a single representation. \item The CNN representation then goes through two Highway layers \cite{srivastava2015training} that apply non-linearities with residual connections before being projected down to a final embedding size which we chose to be coherent with BERT's 768-dimensional wordpiece representations.

\end{enumerate}

As with BERT, we add the token embedding (here, the Character-CNN representation) to position and segment embeddings before feeding the resulting context-independent representation to several Transformer layers. Since CharacterBERT does not split tokens into wordpieces, each input token is assigned a single final contextual representation by the model.

\subsection{Pre-training Procedure}

Like BERT, our model is pre-trained on two tasks: a Masked Language Modelling task (MLM) and a Next Sentence Prediction task (NSP). The only difference lies in the MLM task where instead of predicting single wordpieces, we predict entire words. This natural consequence of handling words instead of wordpieces is somewhat related to recent work on Whole Word Masking which has been shown to improve the quality of BERT models\footnote{Google updated \href{https://github.com/google-research/bert}{their repository} with Whole Word Masking models that improve over the original BERT.} \cite{cui2019pre}.

\section{Experiments}

We compare BERT and CharacterBERT on multiple medical tasks to evaluate the impact of using a Character-CNN module instead of wordpieces. In an attempt to dissociate this impact from any other effects that may be related to training models in our own specific settings, we train each CharacterBERT model alongside a BERT counterpart in the exact same conditions.

\subsection{Model Settings}\label{modelSettings}

We base our models on the ``base-uncased'' version of BERT, which uses 12 Transformer layers with 12 attention heads and produces 768-d representations from uncased texts. This version has $\approx 109.5$M parameters and the corresponding CharacterBERT architecture has $\approx 104.6$M parameters. It is interesting to note that using a Character-CNN actually results in a smaller overall model despite using a seemingly complex character module. This is because BERT's wordpiece matrix has $\approx$~30K~$\times$~\mbox{768-d} vectors while CharacterBERT uses smaller 16-d character embeddings with mostly small-sized CNNs.

We pre-train four different models to simulate the usual situation where BERT is first pre-trained on a general corpus before being re-trained on a set of specialized texts:

\begin{description}
    \item[BERT\textsubscript{general}:] a general-domain model obtained by pre-training BERT on a general corpus. It uses the same architecture and wordpiece vocabulary as BERT~(base,~uncased). \item[CharacterBERT\textsubscript{general}:] a general-domain model obtained by training CharacterBERT on a general corpus. Besides the Character-CNN, it uses the same architecture as BERT\textsubscript{general}. \item[BERT\textsubscript{medical}:] a medical model obtained by re-training BERT\textsubscript{general} on a medical corpus. \item[CharacterBERT\textsubscript{medical}:] a medical model obtained by re-training CharacterBERT\textsubscript{general} on a medical corpus. This is the Character-CNN analog of BERT\textsubscript{medical}.
\end{description}

\subsection{Pre-training Phase}
\subsubsection{Corpora}

The original BERT was pre-trained on English Wikipedia and BooksCorpus \cite{zhu2015aligning}. Since the latter is not publicly available anymore, we replace it with OpenWebText \cite{Gokaslan2019OpenWeb} to train our general-domain models. We also build a specialized corpus from MIMIC-III and PMC OA abstracts to train our medical-domain models (see Table~\ref{pretrainingCorpora}).

\begin{table}[htpb]
\centering
\setlength{\tabcolsep}{8pt}
\resizebox{!}{40pt}{%
\begin{tabular}{clcc}
\toprule
\textbf{Corpus} & \textbf{Composition} & \textbf{\# documents} & \textbf{\# tokens} \\
\midrule
\multirow{2}{*}[-2pt]{General} & Wikipedia (EN) & $5.99 \times 10^{6}$ & $2.14 \times 10^9$ \\
& OpenWebText & $1.56 \times 10^{6}$ & $1.28 \times 10^{9}$ \\
\midrule
\multirow{2}{*}[-2pt]{Medical} & MIMIC-III & $2.09 \times 10^{6}$ & $5.05 \times 10^{8}$ \\
& PMC OA abstracts & $2.33 \times 10^{6}$ & $5.22 \times 10^{8}$ \\
\bottomrule
\end{tabular}
}
\centering\caption{Statistics on pre-training corpora.}\label{pretrainingCorpora}
\end{table}

\subsubsection{Pre-training Setup}\label{sec:pretrainingSetup}

We train each model using $16$ Tesla\,V100-SXM2-16GB GPUs and following the implementation and parameters in the NVIDIA codebase\footnote{More specifically, we adapt \href{https://github.com/NVIDIA/DeepLearningExamples/tree/master/PyTorch/LanguageModeling/BERT}{these scripts} to our needs.}. Each complete pre-training phase consists of two steps:

\begin{description}
    \item[Step 1] \numprint{3,519} updates with a batch size\footnote{We use gradient accumulation for larger batch sizes.} of \numprint{8,192} and a learning rate of $6.10^{-3}$ on sequences of size 128.
    \item[Step 2] 782 updates with a batch size of \numprint{4,096} and a learning rate of $4.10^{-3}$ on sequences of size 512.
\end{description}

\noindent All models are optimized using LAMB \cite{you2019large} with a warm-up rate and weight decay of $0.01$.

\subsection{Evaluation Phase}
\subsubsection{Tasks}
All models are evaluated on five medical tasks after adding task-specific layers \cite{devlin-etal-2019-bert}.

\begin{description}
    \item[Medical Entity Recognition] We evaluate our models on the i2b2/VA 2010 \cite{uzuner20112010} clinical concept extraction task which aims to extract three types of medical concepts: \textsc{problem} (e.g.\ ``headache''), \textsc{treatment} (e.g.\ ``oxycodone''), and \textsc{test} (e.g.\ ``MRI'').

    \item[Natural Language Inference] We also evaluate on the clinical natural language inference task MEDNLI \cite{romanov-shivade-2018-lessons} that aims to classify sentence pairs into three categories: \textsc{contradiction}, \textsc{entailment}, and \textsc{neutral}.

    \item[Relation Classification] For more variety, we also evaluate on two biomedical relation classification tasks: ChemProt \cite{krallinger2017overview} from the \href{https://biocreative.bioinformatics.udel.edu/resources/corpora/chemprot-corpus-biocreative-vi/}{BioCreative VI} challenge and DDI \cite{herrero2013ddi} from \href{https://www.cs.york.ac.uk/semeval-2013/task9/}{SemEval 2013 - Task 9.2}. The goal of ChemProt is to detect and classify chemical-protein interactions as \textsc{activator (cpr:3), inhibitor (cpr:4), agonist (cpr:5), antagonist (cpr:6)}, or \textsc{substrate (cpr:9)}. The goal of DDI is to detect and classify drug-drug interactions into the following categories:  \textsc{advise} (\textsc{ddi}-advise), \textsc{effect} (\textsc{ddi}-effect), \textsc{mechanism} (\textsc{ddi}-mechanism), and \textsc{interaction} (\textsc{ddi}-int). 
    
    \item[Sentence Similarity] Finally, we also evaluate our models on the clinical sentence similarity task ClinicalSTS \cite{wang2018medsts} from BioCreative/OHNLP Challenge 2018, Task~2 \cite{wang2018overview}. The goal here is to produce similarity scores for sentence pairs that correlate with the gold standard.

\end{description}

\begin{figure}[htbp]
\begin{center} 
\includegraphics[height=190pt]{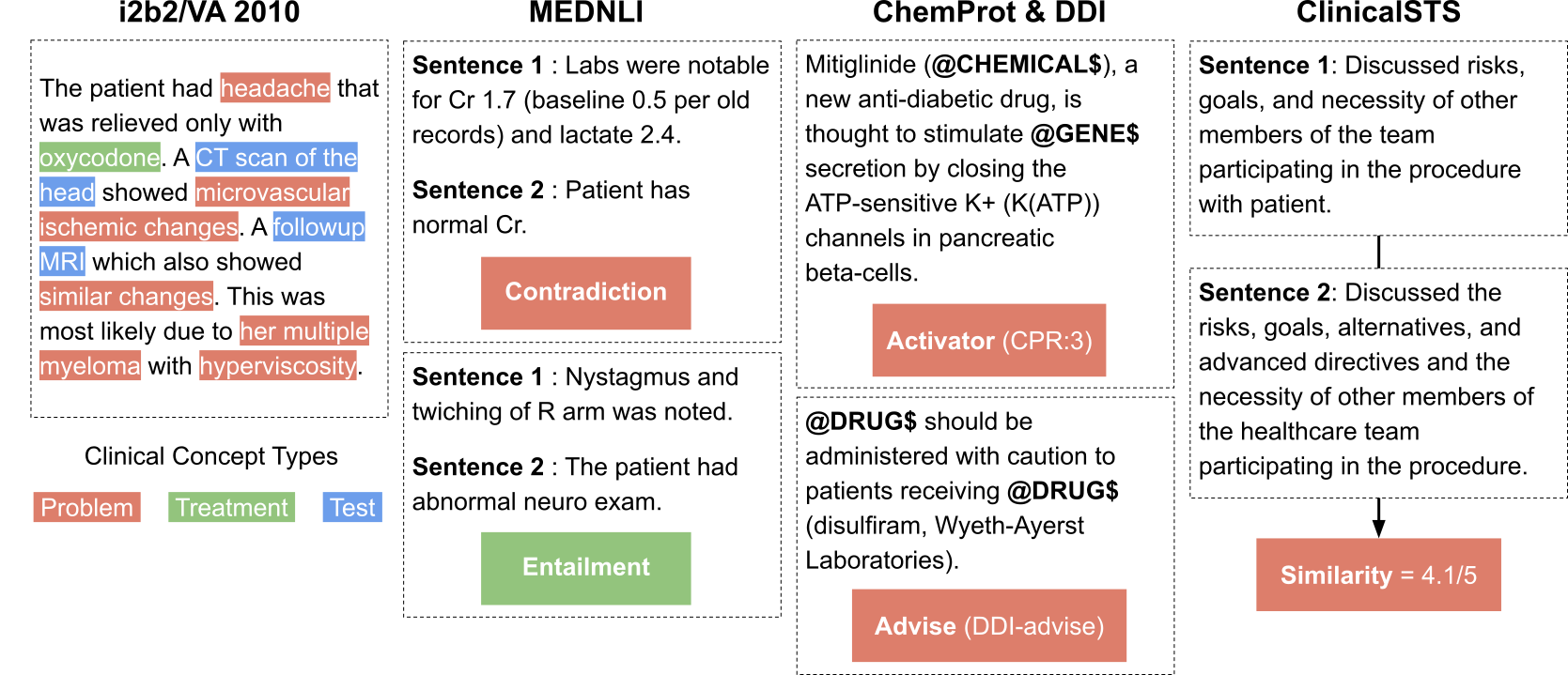}
\end{center} 
\caption{Examples from each evaluation task.} \label{medical_tasks}
\end{figure}

We provide examples for each task in Figure~\ref{medical_tasks} and report the number of examples in Table~\ref{taskStats}.

\begin{table}[htbp]
\centering
\setlength{\tabcolsep}{8pt}
\resizebox{!}{32pt}{%
\begin{tabular}{lccccc}
\toprule
 & \textbf{i2b2} & \textbf{MEDNLI} & \textbf{ChemProt} & \textbf{DDI} & \textbf{ClinicalSTS}\\
\midrule
Train & \numprint{24,757} & \numprint{11,232} & \numprint{19,460} & \numprint{18,779} & 600 \\
Val. & \numprint{6,189} & \numprint{1,395} & \numprint{11,820} & \numprint{7,244} & 150 \\
Test & \numprint{45,404} & \numprint{1,422} & \numprint{16,943} & \numprint{5,761} & 318 \\
\bottomrule
\end{tabular}
}
\centering\caption{Number of examples of each evaluation task.}\label{taskStats}
\end{table}

\subsubsection{Evaluation Setup}

Given all the pre-trained models, the evaluation tasks, and a set of random seeds $i \in {1..10}$:
\begin{enumerate}
    \item We choose a pre-trained model, an evaluation task, and a random seed $i$ then run 15 training epochs with batches of size 32. \item At each epoch, we evaluate the model on a validation set that is either given or computed as 20\% of the training set. According to the validation performance, we save the best model. \item After completing all training epochs, we load the best model and evaluate it on the test set. \item We repeat the whole process for all seeds to compute a final performance as \textit{mean} $\pm$ \textit{std}.
\end{enumerate}

In addition to being useful for measuring model variability, fine-tuning 10 versions for each model also enables us to build ensembles. In fact, by using a majority voting strategy, we are able to combine the predictions from each seed into a single ensemble model\footnote{For ClinicalSTS, we use the average predicted score instead of a majority class since the targets are continuous.}. In practice we do not use all seeds at once: we exclude a single seed, build an ensemble then repeat this process to get 10 ensembles for each model setting which can be used to compute a final ensemble performance as \textit{mean} $\pm$ \textit{std}. All fine-tuning experiments are run on a single Tesla\,V100-PCIE-32GB and are optimized using the Adam optimizer \cite{kingma2014adam} with a learning rate of 3e-5, a warm-up ratio of $0.1$, and a weight decay of $0.1$.

\section{Results and Discussion}

\subsection{Speed Benchmark}
\subsubsection{Pre-training}
Using the setup detailed in Section~\ref{sec:pretrainingSetup}, training a single BERT through Steps~1 and~2 takes around 26.5 hours for BERT and 55 hours for CharacterBERT even though both architectures have about the same number of parameters. This large gap in pre-training speed is partly due to the Character-CNN being slower to train as it is more complex than the original wordpiece embedding matrix. However, the main reason for the slower pre-training is that we are not able to use a very specific trick during Masked Language Modelling. In fact, BERT shares the parameters of its wordpiece embedding matrix with the MLM output layer, which allows it to train faster. In our case, since we do not use wordpieces, we build a temporary vocabulary from the top 100K tokens in the training corpus and use them as targets for MLM\footnote{Please note that this also means that we never mask tokens that are not within the top 100K most frequent tokens.}. We expect that improved pre-training speed can be achieved using Noise Contrastive Estimation \cite{mnih2013learning} or similar methods. However, such optimizations are left for future work.

\subsubsection{Fine-tuning}

In addition to pre-training speed, we also report the fine-tuning speed both at training and inference time.

\begin{figure}[htbp]
\begin{center} 
\includegraphics[height=82pt]{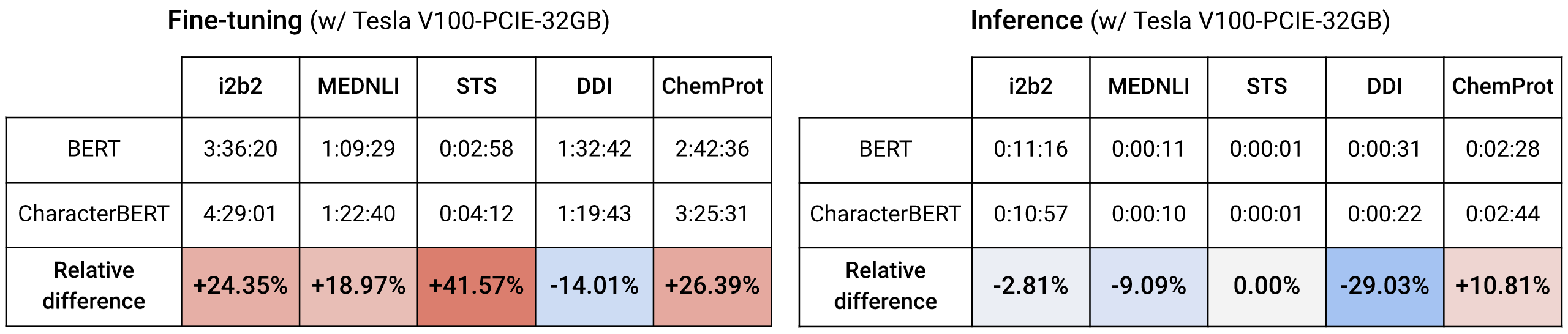}
\end{center} 
\caption{Training/inference speed comparison.} \label{timeDiff}
\end{figure}

Figure~\ref{timeDiff} shows that CharacterBERT is much less at a disadvantage when it comes to fine-tuning (19\% slower on average instead of 108\%). However, in the specific case of the DDI task, CharacterBERT is actually 14\% faster than BERT. This exceptional behavior may be due to the presence of many domain-specific terms that are split into multiple wordpieces, thus increasing the input size with BERT. In fact, since our model works at the word level, the input size is stable and data batches may be processed faster than with BERT. At inference time, CharacterBERT is slightly faster than BERT as the Character-CNN is not as slow during inference as it is during optimization. 

\begin{figure*}[htbp]
\begin{center} 
% 84% is the max before the figure moves to the next page
\includegraphics[width=0.84\textwidth]{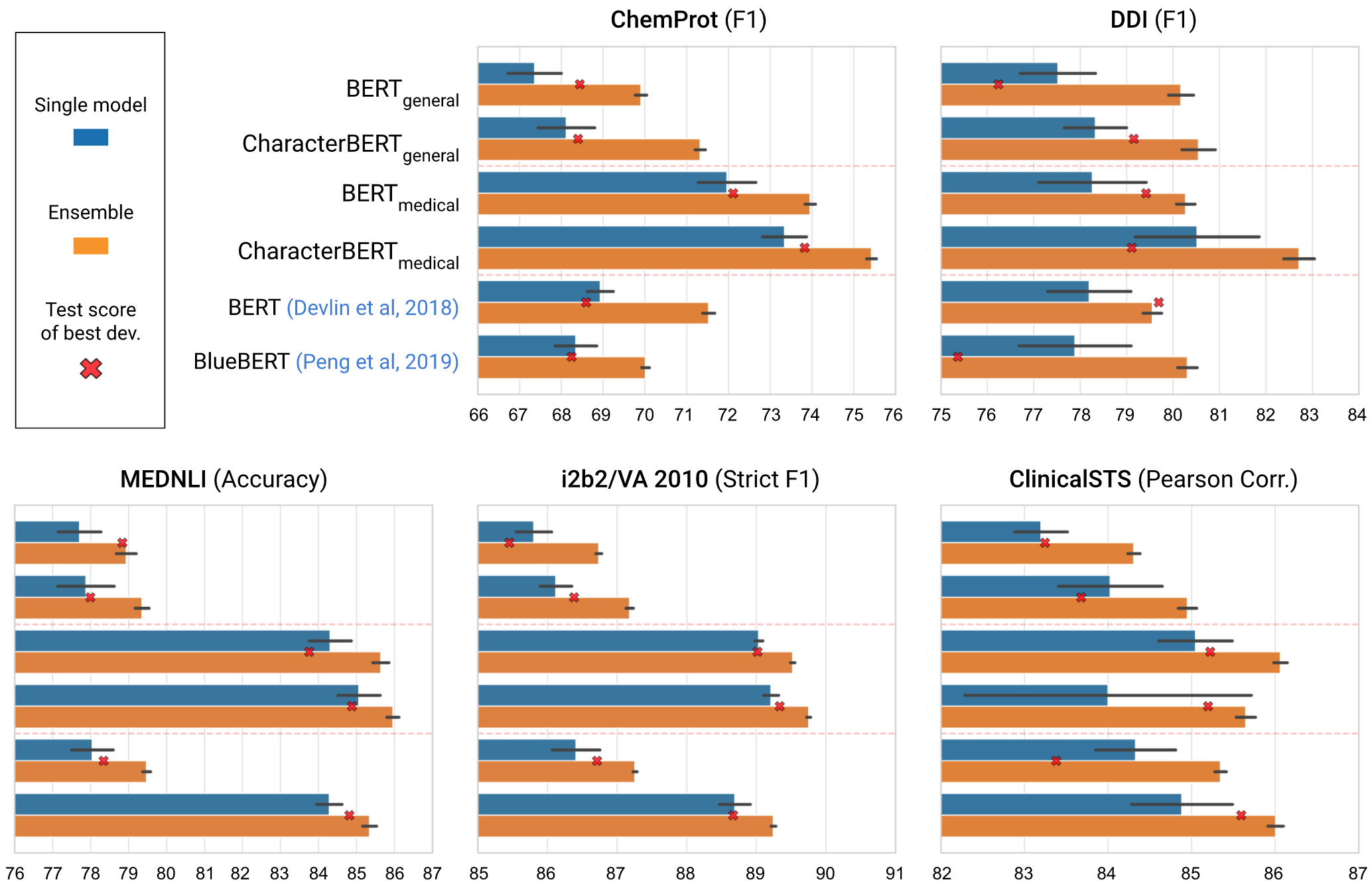}
\end{center}
\caption{Comparison of pre-trained models when fine-tuned on several medical tasks. For each model, the test performance of 10 random seeds is expressed as \textit{mean} $\pm$ \textit{std} and is shown in blue for single models and orange for ensembles. The performance of the best validation seed is shown in red.} \label{results}
\end{figure*}

\subsection{Reproducing Vanilla Models}

We report the performance of BERT(base,~uncased) as well as BlueBERT(base, uncased) \cite{peng2019transfer}, a medical model pre-trained on MIMIC-III and PubMed abstracts\footnote{Note that BlueBERT is trained on PubMed abstracts while our medical models are trained on PMC OA abstracts.}. Including these results allows us to evaluate the quality of our pre-training procedure. Figure~\ref{results} shows that BERT\textsubscript{general} performs slightly worse than the original BERT despite using exactly the same architecture. However, this difference is small and can be attributed to either the different general-domain corpora (OpenWebText instead of BooksCorpus) or to differences in pre-training parameters (number of updates, batch size...). Moreover, we see that BERT\textsubscript{medical} performs at the same level as BlueBERT, sometimes outperforming the latter substantially ($\approx+4$~F1 on ChemProt), which allows us to safely assume our pre-training procedure to be correct.

\subsection{Ensembles and Model Selection}

We can see from Figure~\ref{results} that ensembles (orange bars) clearly improve over single models (blue bars). While not surprising \textit{per se}, it is worth noting that these ensembles were produced using a naive majority voting strategy which can easily be applied as a post-processing step. Moreover, we see that the test results of the best validation model (red symbol) are always below those of the ensembles' performance. Finally, we note that ensembles have substantially lower variances compared to single models, which makes them more reliable for comparisons.

\subsection{BERT vs. CharacterBERT: How Significant Is the Difference?}

\begin{figure*}[h]
\begin{center}
% 76% is the max before the figure moves to the next page
\includegraphics[width=0.76\textwidth]{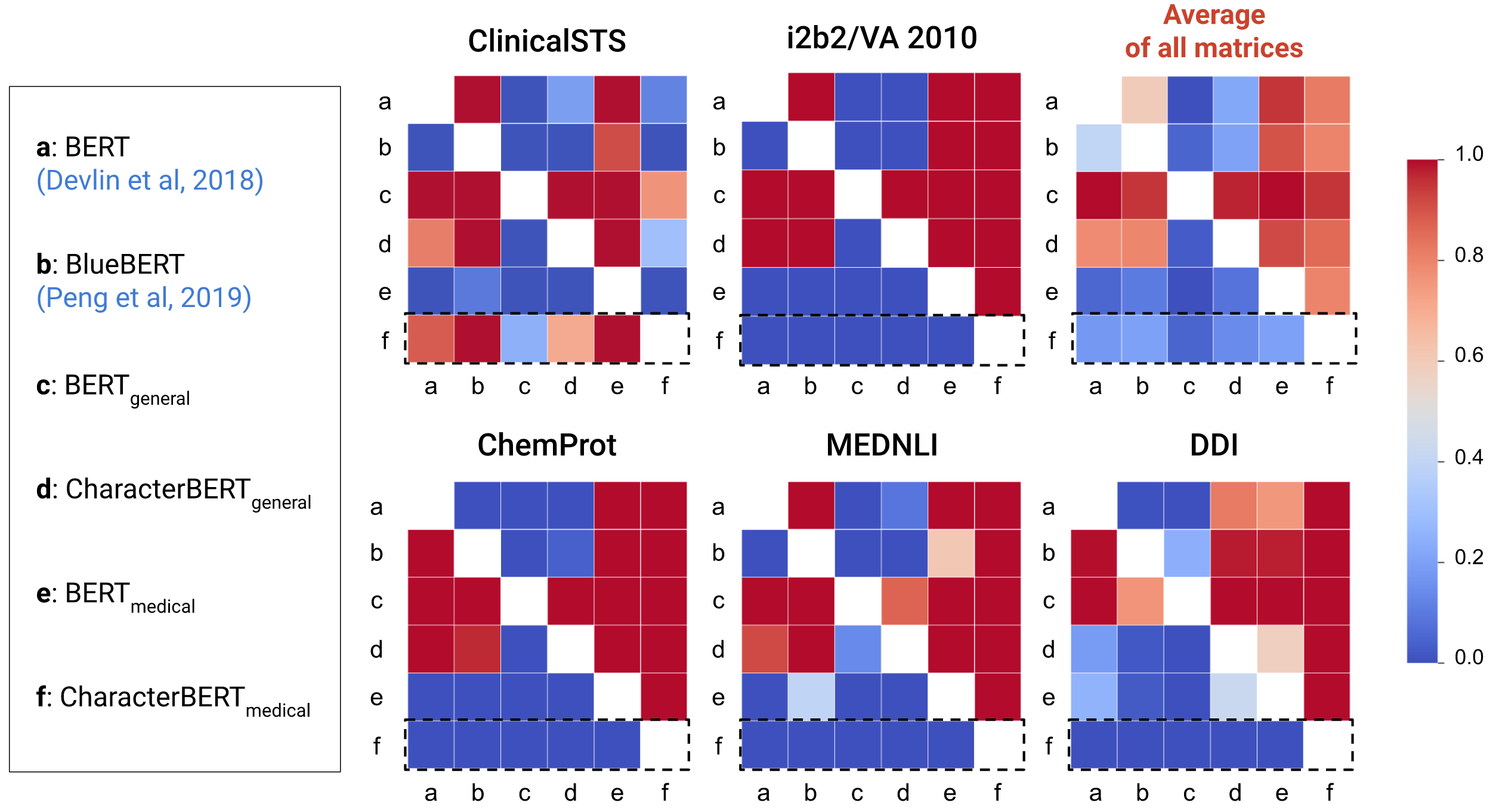}
\end{center} 
\caption{Statistical significance: Minimal distance $\epsilon$ for Almost Stochastic Order at level $\alpha=5\%$. Blue cells mean that the left model is significantly better than the bottom model. Red cells mean the opposite.} \label{ASO}
\end{figure*}

Figure~\ref{results} shows that CharacterBERT often improves over a vanilla BERT. In particular, our medical model improves over the ensemble performance of BERT\textsubscript{medical} by $\approx1.5$ points on ChemProt, $\approx2$ points on DDI, and $\approx0.5$ points on MEDNLI and i2b2. However, we see that CharacterBERT\textsubscript{medical} performs worse than BERT in the specific case of ClinicalSTS and suffers from a surprisingly high variance. Since the ClinicalSTS dataset is also very small compared to the other datasets, these results should be taken with care even if the difference with BERT seems to be significant according to Figure~\ref{ASO}. Results with general-domain models seem to also be in favor of CharacterBERT. However, these differences may not be substantial.

To provide a more rigorous evaluation of the statistical significance of our results, we perform Almost Stochastic Order tests (ASO) \cite{dror-etal-2019-deep} for each pair of models. ASO tests aim to determine whether a stochastic order exists between two models based on their respective sets of evaluation scores. In practice, given the 10 single model scores of two chosen models A and B, the method computes a test-specific value $\epsilon$ that indicates how far model A is from being significantly better than model B. This distance $\epsilon$ is equal to 0 when model A $\succeq$ B, 1 when B $\succeq$ A, and 0.5 when no order can be determined. Figure~\ref{ASO} shows the values of $\epsilon$ for all model pairs on each task. Looking at the average significance matrix, we can see that CharacterBERT\textsubscript{general} improves over its BERT counterpart (cell [d,c]). Moreover, we see that the overall best model is CharacterBERT\textsubscript{medical} as evidenced by the bottom blue row (cells [f,a] to [f,e]), which further validates that our model indeed improves over vanilla BERT.

\subsection{Robustness to Noise and Misspellings}

We want to investigate whether CharacterBERT is more robust to noise and misspellings than BERT. For that purpose, we create noisy versions of the MEDNLI corpus where, given a noise level of X\%, we transform each token with the same probability into a misspelled version either by removing, adding, replacing a single character or swapping two consecutive characters. We conduct experiments where noise is added to the test set only as well as experiments adding noise to the train/dev/test splits.

\begin{figure}[htpb]
\begin{center} 
% 75% is the max before the figure moves to the next page
\includegraphics[width=0.75\textwidth]{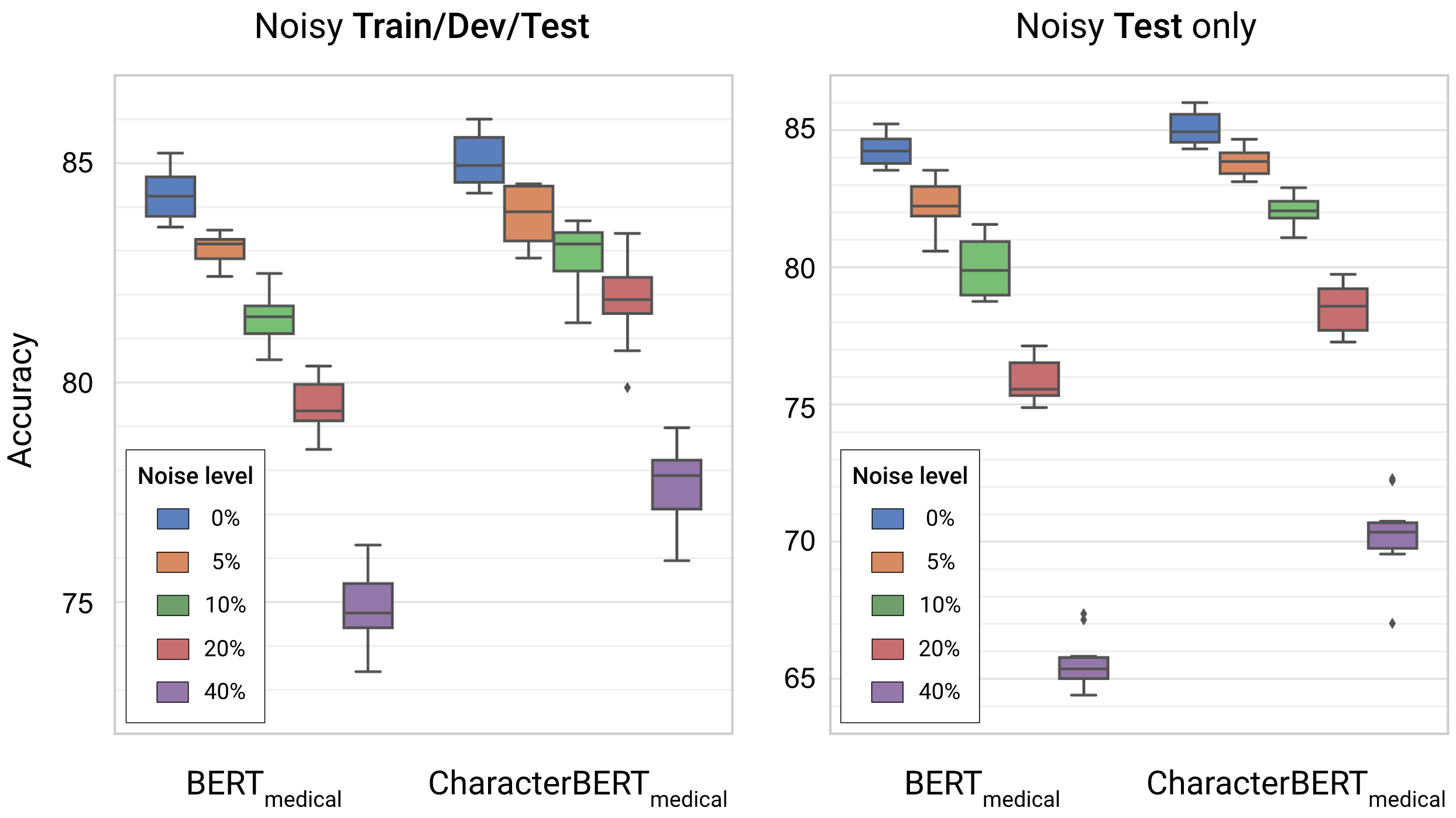}
\end{center} 
\caption{Comparing BERT and CharacterBERT on noisy (misspelled) versions of MEDNLI test.} \label{robustness}
\end{figure}

Figure~\ref{robustness} shows the results for BERT\textsubscript{medical} and CharacterBERT\textsubscript{medical} with various noise levels. We see that the latter is indeed more robust to misspellings as evidenced by the slower decrease in performance compared to BERT. In particular, when a noise level of 40\% is applied to the test set only, CharacterBERT is $\approx 5$ F1 higher than BERT whereas the original difference between the two models was $< 1$ F1. Experiments adding noise to all splits show that both models can learn to be more robust, however, CharacterBERT remains at an advantage.

\subsection{Discussion and Future Work}

Overall CharacterBERT seems to either perform at the same level or improve over BERT. This is especially true for the specialized versions and is further validated by the ASO tests. The new variant also seems to be more robust to misspellings while at the same time producing word-level open-vocabulary representations. This improved robustness is desirable since BERT seems to be sensitive to misspellings \cite{pruthi-etal-2019-combating,sun2020adv}. On the downside, CharacterBERT is slower to pre-train, although not as slow to fine-tune and even slightly faster at inference time. Future work may apply a Character-CNN to recent Transformer-based models \cite{lan2019albert,sun2019ernie}, optimize the pre-training architecture to improve its speed, or explore any other advantages of a character-level system over wordpieces.

\section{Conclusion}

The overall strategy when building specialized versions of BERT seems to be re-training the original model on a specialized corpus. This implies keeping a general-domain wordpiece vocabulary that may not be suited for the domain of interest. Our main contribution is CharacterBERT, a variant of BERT that drops the wordpiece system altogether in favor of a Character-CNN. This module represents tokens by consulting their characters, allowing our model to produce word-level open-vocabulary representations. We evaluate CharacterBERT and show that it globally outperforms BERT when specialized for the medical domain while at the same time being more robust to misspellings.

\section*{Acknowledgments}
This  work  has  been  funded  by  the  French  National Research Agency (ANR) and is under the ADDICTE project (ANR-17-CE23-0001). Moreover, computational resource of AI Bridging Cloud Infrastructure (ABCI)\footnote{\url{https://abci.ai/}} provided by National Institute of Advanced Industrial Science and Technology (AIST) was used.

\bibliographystyle{coling}
\bibliography{coling2020}

\clearpage

\appendix
\section{Appendices}
\label{sec:appendices}

\subsection{Detailed Test Scores}
\begin{figure}[htbp]
\begin{center} 
\includegraphics[width=\textwidth]{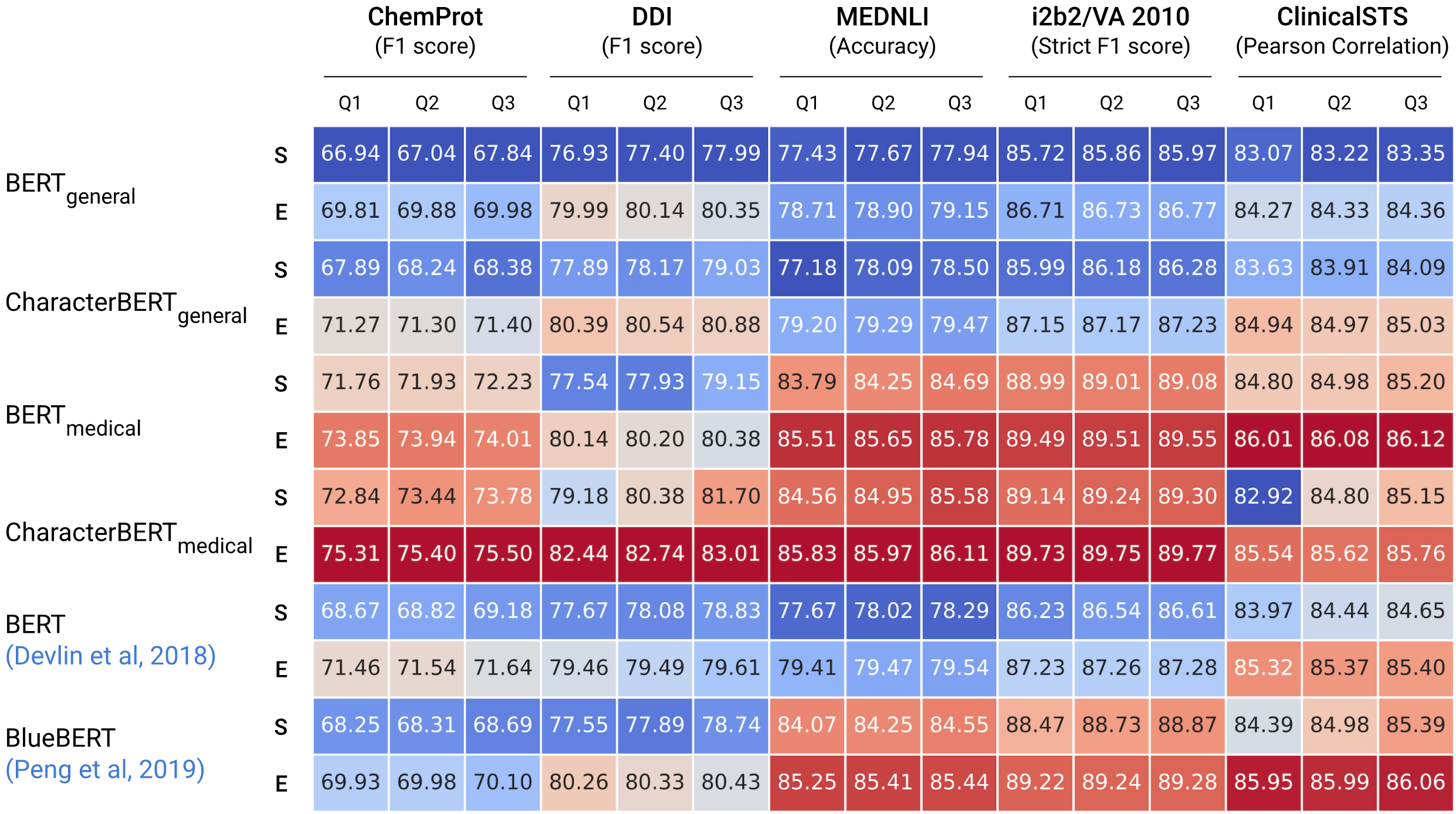}
\end{center} 
\caption{Performance of our pre-trained models when fine-tuned on five different medical tasks. Two baselines are included: BERT \protect\cite{devlin-etal-2019-bert} using the ``base-uncased'' architecture, and BlueBERT \protect\cite{peng2019transfer} a medical BERT that is the result of re-training the former model on MIMIC-III and PubMed abstracts. Legend: Qi = i-th quartile, E = ensemble, S = single model.} \label{results_table}
\end{figure}

Figure~\ref{results_table} provides numerical evaluation scores for the models displayed in Figure~\ref{results}. To give a better idea about the distribution of model scores, these are reported as first, second (median), and third quartiles.

\end{document}